# Sample-and-Accumulate Algorithms for Belief Updating in Bayes Networks

**Eugene Santos Jr.**
Dept. of Elec. and Comp. Eng.
Air Force Inst. of Tech.
Wright-Patterson AFB
OH 45433-7765
esantos@afit.af.mil

**Solomon Eyal Shimony**
Dept. of Math. and Comp. Sci.
Ben Gurion University of the Negev
P. O. Box 653
84105 Beer-Sheva, Israel
shimony@cs.bgu.ac.il

**Edward Williams**
Dept. of Elec. and Comp. Eng.
Air Force Inst. of Tech.
Wright-Patterson AFB
OH 45433-7765
ewilliam@afit.af.mil

## Abstract

Belief updating in Bayes nets, a well known computationally hard problem, has recently been approximated by several deterministic algorithms, and by various randomized approximation algorithms. Deterministic algorithms usually provide probability bounds, but have an exponential runtime. Some randomized schemes have a polynomial runtime, but provide only probability estimates.

We present randomized algorithms that enumerate high-probability *partial* instantiations, resulting in probability bounds. Some of these algorithms are also sampling algorithms. Specifically, we introduce and evaluate a variant of backward sampling, both as a sampling algorithm and as a randomized enumeration algorithm. We also relax the implicit assumption used by both sampling and accumulation algorithms, that query nodes must be instantiated in all the samples.

## 1 INTRODUCTION

Computing marginal probabilities in a multiply connected Bayes network (also called belief updating [Pearl, 1988]) is an important issue in probabilistic reasoning. The problem is known to be NP-hard [Cooper, 1990], and in fact even approximating the probabilities was shown to be NP-hard [Dagum and Luby, 1993]. Nevertheless, a large number of algorithms addressing the problem of inference in Bayesian networks exist, roughly categorized into exact algorithms, and approximation algorithms. There is quite a large number of exact algorithms and their variations. All of the algorithms have an exponential runtime, where the term in the exponent is some function of the topology.

The class of approximation algorithms can be subclassified into deterministic and randomized algorithms. Most deterministic schemes are base on (partial) enumeration of an exponential number of instantiations (also called assignments), terms, or other aspects of the distribution. By considering these elements starting from the most probable ones, and computing their cumulative probability mass, these algorithms get a successively better approximation as more processing is performed, as follows. Let $\mathcal{E}$ be the evidence, and $q$ be a query node with states $D_q$ (domain of $q$). Let $q_i$ be the ith state of $q$, with $1 \le i \le |D_q|$. Let $\mathcal{Q}_i$ denote the assignment $\{q = q_i\}$. Define the following quantities:

$$\hat{P}(\mathcal{E}) = P(\text{union of elements consistent w. } \mathcal{E}) \quad (1)$$

$$\hat{P}(\neg \mathcal{E}) = P(\text{union of elem. inconsistent w. } \mathcal{E}) \quad (2)$$

$$\hat{P}(\mathcal{E}, \mathcal{Q}_i) = P(\text{elem. contained in } \mathcal{E} \text{ and } \mathcal{Q}_i) \quad (3)$$

$$\epsilon = 1 - \hat{P}(\neg \mathcal{E}) - \hat{P}(\mathcal{E}) \quad (4)$$

where "elements" stands for "events corresponding to already enumerated instantiations" (or terms). The term "inconsistent" above means that the probability of the event intersection is 0, while "contained in" is in the sense of set inclusion of events (a condition stronger than "consistent"). These conditions, in effect, require that the following assumption holds:

**Assumption 1** *In each element, node $q$ must be instantiated.*

With the above definitions, we get bounds on the marginal posterior probabilities (adapted from [Poole, 1993b; Santos and Shimony, 1994]):

$$\frac{\hat{P}(\mathcal{E}, \mathcal{Q}_i) + \epsilon}{\hat{P}(\mathcal{E}) + \epsilon} \ge P(\mathcal{Q}_i \mid \mathcal{E}) \ge \frac{\hat{P}(\mathcal{E}, \mathcal{Q}_i)}{\hat{P}(\mathcal{E}) + \epsilon} \quad (5)$$

and the error margin in the posterior probability, for any $P(\mathcal{Q}_i \mid \mathcal{E})$, is thus: $\Delta = \frac{\epsilon}{\hat{P}(\mathcal{E})+\epsilon}$.

Most such algorithms can provide guaranteed bounds similar to the above on the error of their probability estimates, and if allowed an exponential computation time (which is rarely done for these algorithms), will eventually enumerate all the elements and give an exact probability. We note in passing that the above equations can be used to approximate general distributions, regardless of whether they are represented as Bayes nets. The runtime and quality of approximation of these algorithms usually depends on the actual conditional distributions in the network, rather than just on the topology.

Various algorithms of this class exist. Bounded conditioning [Horvitz et al., 1989] is based on cutset conditioning, but does not sum up the probabilities computed for *all* possible instantiations of the cutset variables, instead starting the evaluation with the most probable instantiations. Algorithms that simply enumerate instantiations are presented in [Poole, 1993b] (enumeration of complete instantiations) and in [Santos and Shimony, 1994] (partial, IB assignments). Another such algorithm considers terms, rather than instantiations [Li and D'Ambrosio, 1992]. Deterministic approximation algorithms that do not fit into this pattern are [Wellman and Liu, 1994; Kjaerulff, 1994].

The above approximation algorithms perform better if the conditional distributions are heavily skewed [D'Ambrosio, 1993; Poole, 1993b][1]. Encouraging theoretical results presented in [Druzdzel, 1994] state that even for weak skewness, a small fraction of the instantiations is expected to hold most of the probability mass. Nevertheless, finding these high-probability instantiations is a hard problem in and of itself.

Randomized approximation algorithms usually depend on some form of sampling or scoring, over a large number of random trials. The probability of an event is estimated based on the fraction of the trials in which the event appears, among the total number of trials. In [Henrion, 1988], approximation is achieved by stochastically sampling instantiations of the network variables. Later work in randomized approximation algorithms attempts to increase sampling efficiency [Bouckaert, 1994; Dagum and Chavez, 1993], and to handle the case where the probability of the evidence is very low [Fung and Favero, 1994], which is a serious problem for most sampling algorithms. The randomized approximation algorithms perform better if the distributions are nearly uniform. In [Dagum and Chavez, 1993], an explicit bound on the runtime is made in terms of a dependence value, which tends to 1 as the conditional probabilities for each node approach uniformity. However, regardless of the exact method employed, these algorithms can only provide either estimates on the errors of their answers, or bounds correct with a certain probability.

This paper aims to take advantage of the randomization (in the search for high-probability instantiations), without losing the guaranteed error bounds provided by the deterministic algorithms. The basic idea is to find the high-probability instantiations with a randomized algorithm, and then to take the cumulative mass in the high-probability instantiations into account when approximating the marginal probability. A drawback is that even with the results of [Druzdzel, 1994], a small fraction of the instantiations is still prohibitively large. It should be possible to use the topology *and* the structure of the local conditional distribution (exhibited in nodes such as noisy OR) to accumulate elements with still higher mass per element.

In previous work [Santos and Shimony, 1994], we presented deterministic algorithms that enumerate Independence-Based (IB) assignments, and accumulate their mass to approximate the marginal probabilities. The algorithms consist of a generator, that provides IB assignments in decreasing order of probability, and an evaluator, that accumulates the mass in the assignments and computes the probability estimates and error bounds. Three generators were examined and evaluated empirically: simple heuristic search, heuristic search with cost-sharing, and integer linear programming. The latter generators allowed the overall algorithms to operate efficiently, comparing favorably with stochastic simulation. The algorithm provided approximations for problem instances that exact algorithms could not handle, on belief networks with 50 nodes or more.

### 1.1 DEFINITIONS AND NOTATION

For convenience, we define our notation, and review the definition of IB assignments, below. An assignment $\mathcal{A}$ is an instantiation to a set of network variables, denoted by a set of (node, state) pairs, or a set of node=state assignments; its set of assigned nodes is denoted span($\mathcal{A}$). $\mathcal{A}$ is complete w.r.t. a node set $S$ just when $S = \text{span}(\mathcal{A})$, and is partial otherwise. $\mathcal{A}$ is consistent if each node appears in at most one pair (that is, if $\mathcal{A}$ is a partial function form a node to a state). Two assignments are consistent just when their union is consistent. The event corresponding to an assignment $\mathcal{A} = \{(v_1, s_1), ..., (v_n, s_n)\}$ is the event where $v_i$ is in state $s_i$, for all $1 \leq i \leq n$. We use $P(\mathcal{A})$ to denote the probability of the event corresponding to $\mathcal{A}$. The term $\mathcal{A}(v)$ denotes the value assigned to $v$ by assignment $\mathcal{A}$. For a node $v$, we denote its parents (immediate predecessors) by $\pi(v)$, all its ancestors (excluding $v$) by $\pi^+(v)$ ($\pi^+$ is the transitive closure of $\pi$), and its ancestors (inclusive of $v$) by $\pi^*(v)$ (reflexive, transitive closure of $\pi$).

A (possibly partial) assignment $\mathcal{A}$ is IB if for every node $v$ instantiated in $\mathcal{A}$, the IB condition holds at $v$ w.r.t. $\mathcal{A}$, where:

**Definition 1** *The IB condition holds at node $v$ w.r.t. $\mathcal{A}$ iff[2] $P(\mathcal{A}_{\{v\}}|\mathcal{A}_{\pi(v)})$ is independent of every possible instantiation of nodes in $\pi^+(v)$, which are not instantiated in $\mathcal{A}$.*

---

[1] Oddly enough, since [Dagum and Luby, 1993] show that a high dependence value (a notion similar to skewness) makes belief updating *NP-hard*. Nevertheless, if we can find the high-probability elements, which is usually the case in practice, it is better if the distribution is skewed.

[2] We use the notation $\mathcal{A}_S$ to denote $\mathcal{A} - \{(v, d) \mid v \notin S\}$ for any set of nodes $S$.

For example, for binary OR node $v$ with binary parents $u_i$, let $\mathcal{A} = \{v = T, u_1 = T\}$. Then the IB condition holds at $v$ w.r.t. $\mathcal{A}$ because $P(v = T \mid u_1 = T) = 1$, and will stay 1, no matter what further instantiations we make to ancestors of $v$: that is, $P(v = T \mid u_1 = T, u_i = T) = P(v = T \mid u_1 = T, u_i = F) = 1$ for any $i > 1$, and this also holds for any set of (possibly indirect) ancestors of $v$ that does *not* include $u_1$.

A natural unit to use for IB assignments is the *maximal IB hypercube* [Santos and Shimony, 1994], defined as follows. A *hypercube* $\mathcal{H}$ is an assignment to a node $v$ and some of its parents. (We say that such a hypercube $\mathcal{H}$ is *based* on $v$). $\mathcal{H}$ is an IB hypercube if the IB condition holds at $v$ w.r.t. $\mathcal{H}$. It is a *maximal* IB hypercube if there is no (set-wise) smaller assignment such that the IB condition still holds at $v$.[3] Clearly, the maximal IB hypercube is not unique. For example, let $v$ be an OR node with parents $u_1, u_2$. Then $H_1 = \{v = T, u_1 = T\}$, $H_2 = \{v = T, u_2 = T\}$, and $H_3 = \{v = T, u_1 = F, u_2 = F\}$ are all maximal IB hypercubes. For each hypercube $H$, we define a *hypercube probability* $P'(H)$ as its conditional probability (rather than the probability of the assignment $H$). In the above example, $P'(H_1)$ is $P(v = T|u_1 = T)$, which is equal to both $P(v = T|u_1 = T, u_2 = T)$, and $P(v = T|u_1 = T, u_2 = F)$, by definition of IB hypercubes. The latter two numbers appear in the distribution array for node $v$ in the Bayes network.

Every IB assignment can be (efficiently) segmented into (possibly overlapping) maximal IB hypercube components. The probability of the assignment is equal to the product of its component hypercube probabilities (the segmentation is not unique, but this holds for all of them). Hypercubes are indeed used as basic elements in several of our approximation algorithms.

## 1.2 PROBLEMS ADDRESSED HERE

The fact that in an IB assignment, not all variables are instantiated, leads to a possible overlap between the events corresponding to different IB assignments. Finding the most probable IB assignment is (in practice) somewhat easier than finding the MAP (most probable complete assignment). Computing the probability of an IB assignment takes roughly linear time in the cardinality of the assignment.

Nevertheless, clearly there exist problem instances for which these generators will not provide even the first most-probable assignment in reasonable time, as this is also an NP-hard problem [Shimony, 1994]. In this paper, we replace the generator with a randomized search algorithm. This entails several complications: first, we cannot be sure when we get the most-probable assignment, let alone use the algorithm to enumerate them in order of decreasing probabilities. However, the evaluator component uses an equation (a variant of equation 5) that assumes nothing about the order of the assignments it processes. It is sufficient that we get the high probability instantiations efficiently, independent of the order. In fact, the overlap (in terms of sample space events) between IB assignments makes it possible to compute a good approximation without ever encountering the most probable assignments.

The overlapping IB assignments lead to the second complication, which is that if there are too many overlaps, computing the cumulative mass (e.g. by inclusion-exclusion) becomes difficult. Early experiments on the behavior of inclusion-exclusion on sets of IB assignments generated using the our deterministic algorithms, showed that the problem is benign in practice. Nevertheless, if we now relax the requirement that the assignments arrive in decreasing order of probability, we should consider the possibility that the behavior of inclusion-exclusion deteriorate.

Randomized search for a good set of IB assignments is possible in various ways. In fact, one could simply take any algorithm in the class of randomized approximation algorithms, and score the generated instantiations as in equation 5, and in this manner get both the hard bounds and, if they are unreasonable, use the probability estimate from sampling instead. In this paper, we introduce a novel variant of the backward simulation algorithm [Fung and Favero, 1994], and show its convergence to the correct values (theorem 1). Its advantage is that sampling larger chunks of the probability space is likely to converge faster than sampling complete instantiations, which is done in most sampling algorithms. In addition to simulation, the algorithm also enumerates the sampled IB assignments. We also experiment with genetic algorithms as a possible source of good IB assignments.

Since IB assignments do not instantiate all nodes, it seems that such algorithms must be explicitly required to instantiate each query node, in order to comply with assumption 1, at a considerable increase in computation time. Essentially, this implies that the algorithm must be re-run once for every query node. We show that under certain conditions, assumption 1 can be significantly relaxed, obviating the need for multiple runs of the algorithm, for an interesting class of problems. The latter is a significant theoretical (and practical) extension of the results of [Santos and Shimony, 1994].

The rest of the paper is organized as follows. In section 2, we outline the design of the randomized part of the algorithm, and the evaluator. Section 3 shows how to sample IB assignments, and how to relax assumption 1, both for accumulation and for sampling. Section 5 is an empirical evaluation of the algorithm and a performance comparison to other algorithms.

## 2 HYBRID ALGORITHM

The algorithm consists of an instantiation generator (the randomized algorithm), which outputs IB assign-ments to a summation evaluator and to a sampling evaluator. The summation evaluator we used initially is that of [Santos and Shimony, 1994]. This simplified evaluator puts the instantiation given by the generator into the ith bucket (for an IB assignment consistent with the evidence and $\mathcal{Q}_i$) or to the 0th bucket (for an IB assignment inconsistent with the evidence). Instantiations that are completely subsumed by previous ones are discarded. The effective mass of the assignment (probability of the assignment that is not overlapped by any assignment already in the bucket) is added to $\dot{P}(\mathcal{E}, \mathcal{Q}_i)$, or $\dot{P}(\neg\mathcal{E})$ respectively.

---

[3]Setwise smaller assignments "cover" *larger* parts of the sample-space, hence the term "maximal IB hypercube".

The sampling evaluator is the likelihood weighing scoring method [Henrion, 1988], that scores each IB assignment $\mathcal{A}$ according to its sampling probability $P_S(\mathcal{A})$ and its event probability $P(\mathcal{A})$. That is, for each query node $q$, a set of total scores $s_t(q_i)$ (one for each state, initialized to 0) is kept. If $\mathcal{A}(q) = q_i$, it increments the score for state $i$ by $s_\mathcal{A}(q_i) = \frac{P(\mathcal{A})}{P_S(\mathcal{A})}$. The probability estimate $\hat{P}(\mathcal{Q}_i \mid \mathcal{E})$ is given by:

$$\hat{P}(\mathcal{Q}_i \mid \mathcal{E}) = \frac{s_t(q_i)}{\sum_{q_j \in D_q} s_t(q_j)}$$

Obviously, the sampling evaluator does not need to keep track of already visited instantiations.

In the generator part, we need to be able to provide the high-probability instantiations quickly. Since we do not need to do that in strict order, any random walk that visits the high-probability instantiations frequently is a viable choice. In order to use the sampling scorer, it is necessary, in addition to the random walk, that $P_S(\mathcal{A})$ be known.

## 3 SAMPLING IB ASSIGNMENTS

In order to be able to perform a sampling algorithm, we need to be able to compute a meaningful sampling probability of an instantiation, $P_S(\mathcal{A})$. The fact that maximal IB hypercubes based on a variable may be overlapping events (as in the example in section 1) makes this difficult. Things become easier if we only consider a set of disjoint (that is, mutually inconsistent) IB instantiations that covers the probability space. In order to do that, we use, at each node $v$, a set of disjoint, covering, IB hypercubes $\mathcal{H}'_v$, instead of the set of maximal hypercubes. In the example, we could use the set $\{H'_1, H'_2, H'_3\}$, with $H'_1 = \{v = T, u_1 = T\}$, $H'_2 = \{v = T, u_1 = F, u_2 = T\}$, and $H'_3 = \{v = T, u_1 = F, u_2 = F\}$, rather than the non-disjoint set $\{H_1, H_2, H_3\}$.

### 3.1 THE BASIC SAMPLING ALGORITHM

Now, let us define the following IB assignment selection method (METHOD 1).

**Input:** Bayes network with a set $\mathcal{H}'_v$ of disjoint covering hypercubes for each node, an evidence assignment $\mathcal{E}$, and (optionally) a query node $q$.

**output:** An assignment $\mathcal{A}$.

1. Order the nodes in a reverse topological ordering (where nodes precede their parents). Discard every node preceding all evidence and query nodes.
2. Let $\mathcal{A} = \mathcal{E}$, and mark all nodes as unvisited.
3. OPTION 1: Select a state $q_i$ for the query node, and set $\mathcal{A}$ to $\mathcal{A} \cup \mathcal{Q}_i$.
4. Visit (and mark as visited) the first unvisited node $v$ in span($\mathcal{A}$), according to the ordering, and:
   (a) Select a hypercube $H_v$ consistent with $\mathcal{A}$ from the set $\mathcal{H}'_v$.
   (b) Set $\mathcal{A}$ to $\mathcal{A} \cup H$.
5. Repeat step 4 until no unvisited nodes remain in span($\mathcal{A}$).

The selection in steps 2 and 4a are deliberately left arbitrary at this point, to be determined by context below (heuristic, non-deterministic, or randomized).

Let $\tilde{\mathcal{A}}$ denote the set of all assignments that can be (non-deterministically) generated by METHOD 1. Let $\tilde{\mathcal{A}}^{q_i}$ be the set of all IB assignments $\mathcal{A}$ in $\tilde{\mathcal{A}}$ such that $\mathcal{Q}_i \subseteq \mathcal{A}$, and $\tilde{\mathcal{A}}^q$ be the set of all IB assignments in $\tilde{\mathcal{A}}$ that assign *some* value to $q$. We can now show:

**Lemma 1** *Let $\mathcal{A} \in \tilde{\mathcal{A}}$. Then $\mathcal{A}$ is an IB assignment, and for each $v \in span(\mathcal{A})$, there is a unique $H_v \in \mathcal{H}'_v$ such that $H_v \subseteq \mathcal{A}$.*

Proof: Since $\mathcal{A}$ is a union of consistent IB hypercubes, and every node in span($\mathcal{A}$) is visited, then the IB condition holds at every node in span($\mathcal{A}$), and thus $\mathcal{A}$ is by definition an IB assignment. Selection of the hypercubes at each node $v$ is unique, because only one hypercube in $\mathcal{H}'_v$ is consistent with $\mathcal{A}$ (disjointness of the hypercubes in $\mathcal{H}'_v$). □

**Lemma 2** *The set of IB assignments $\tilde{\mathcal{A}}$ is disjoint, and covers the evidence.*

Proof outline: Let $\mathcal{B}$ be any complete assignment to the nodes of the diagram, consistent with the evidence. We show that there is some $\mathcal{A} \in \tilde{\mathcal{A}}$ such that $\mathcal{A} \subseteq \mathcal{B}$, i.e. the event $\mathcal{B}$ is a sub-event of $\mathcal{A}$, by tracing METHOD 1. Disjointness follows from lemma 1.

METHOD 1 (including OPTION 1), and lemmas 1,2, allow us to define the random selection of IB assignments as follows. To generate a random IB instantiation, use METHOD 1, but in step 3 select state $q_i$ randomly with some probability $P_S(q_i)$, and in step 4a independently select hypercube $H_v$ with some probability $P_S(H_v)$. The only constraints on the selection probabilities are that in step 3, the $q_i$s are selected with strictly positive probabilities that sum to 1, that at each node $v$ visited the probability of selecting some hypercube $H_v$ is 1, and that $P_S(H_v) > 0$ for all $H_v \in \mathcal{H}_v$. Let $H_{span(\mathcal{A})}$ be the set of hypercubes selected in generating assignment $\mathcal{A}$.

**Proposition 1** *Let $\mathcal{A}$ be any IB assignment that can be generated by the random selection method above. Then the probability of generating $\mathcal{A}$ (the sampling probability) is given by:*

$$P_S(\mathcal{A}) = P_S(q_i) \prod_{H \in H_{span(\mathcal{A})}} P_S(H)$$

The above follows immediately from the selection method, the selections being performed independently, and lemma 1. Finally, we can show that the random selection method above, together with the sample scoring, constitute a valid approximation algorithm.

**Theorem 1** *The sampling algorithm using the random selection method, and likelihood weighing scoring, converges to the correct value of $P(Q_i \mid \mathcal{E})$.*

Proof outline: It is sufficient to show that, for a single sample, the expected value of the sample score is equal to $P(\mathcal{E} \cup Q_i)$, which follows from proposition 1, and lemma 2. The theorem follows immediately from prior work [Fung and Favero, 1994].

### 3.2 RELAXING ASSUMPTION 1

We now relax the requirement that the query node be instantiated, and require instead only that $q$ be either instantiated, or independent of the sampled IB assignments. We begin by modifying the requirement for accumulation. Let $\tilde{\mathcal{A}}^{q'}$ be all the IB assignments $\mathcal{A}$ in $\tilde{\mathcal{A}}$ such that $\pi^*(q) \not\subseteq span(\mathcal{A})$ (that is, assignments that do not assign either $q$ or any of its ancestors).

**Theorem 2** *Let $\tilde{\mathcal{A}}$ be the set of IB assignments generated non-deterministically by METHOD 1 without OPTION 1, and $q$ be an arbitrary node. If $\tilde{\mathcal{A}} = \tilde{\mathcal{A}}^{q'} \cup \tilde{\mathcal{A}}^q$, then:*

$$P(\mathcal{E} \cup Q_i) = \sum_{\mathcal{A} \in \tilde{\mathcal{A}}^{q_i}} P(\mathcal{A}) + \sum_{\mathcal{A} \in \tilde{\mathcal{A}}^{q'}} P(\mathcal{A})P(Q_i) \quad (6)$$

Proof: We begin by showing the lemma:

**Lemma 3** *If $q \cup \pi^+(q) \not\subseteq span(\mathcal{A})$, then $P(\mathcal{A} \cup Q_i) = P(\mathcal{A})P(Q_i)$ for every $q_i \in D_q$.*

The theorem follows immediately from lemma 3, $\tilde{\mathcal{A}}^{q'} \cup \tilde{\mathcal{A}}^{q_i}$ being a disjoint set of IB assignments covering $\mathcal{E}$, and if $j \neq i$, then $P(\mathcal{A} \cup Q_i) = 0$ for all $\mathcal{A} \in \tilde{\mathcal{A}}^{q^j}$. □

**Corollary 1** *Equation 6 always holds whenever $q$ is a root node.*

The best-first approximation algorithm of [Santos and Shimony, 1994] essentially used METHOD 1 for generating IB assignments, except that an agenda of assignments was kept, and the selection at step 4 is done on the "best" instantiation in the agenda, using a heuristic (both "cost-so-far" and "shared-cost" were tried). The selection then selected "in parallel" all possible hypercubes in step 4a, and all the resulting assignments were put on the agenda. Additionally, METHOD 1 was implemented as a generator: that is, after returning an IB assignment, it was *resumed* at step 4. In the algorithm, OPTION 1, instantiating the query node, was always used.

However, with the above corollary, whenever we want to approximate probabilities only for root nodes, we need not force the nodes to be instantiated in the IB assignments, i.e. we can drop OPTION 1. Additionally, finding the prior probability of each root node takes time $O(1)$. In fact, even for non-root nodes, OPTION 1 is not necessary as an initializing step for an accumulation-type algorithm. Use METHOD 1 (without OPTION 1) to find an IB assignment $\mathcal{A}$. Now, if $q$ is instantiated in $\mathcal{A}$ or no ancestor of $q$ is in $\mathcal{A}$ (which is sure to occur if $q$ is a root node), we are done (assuming the prior probabilities for $q$ are known). Otherwise, add $(q, q_i)$ to $\mathcal{A}$, and continue the hypercube selection process until termination. If that is done, one can still use equation 6.

It is interesting that relaxing the requirement of instantiated query nodes can also be used in sampling.

**Theorem 3** *Let $\tilde{\mathcal{A}}$ be the set of IB assignments generated non-deterministically by METHOD 1 without OPTION 1, and $q$ be an arbitrary node. If $\tilde{\mathcal{A}} = \tilde{\mathcal{A}}^{q'} \cup \tilde{\mathcal{A}}^q$, and the following sampling weight is used:*

$$s_\mathcal{A}(q_i) = \frac{P(\mathcal{A})}{P_S(\mathcal{A})} \begin{cases} 1 & Q_i \in \mathcal{A} \\ P(Q_i) & q \notin span(\mathcal{A}) \\ 0 & otherwise \end{cases} \quad (7)$$

*then, using METHOD 1 without OPTION 1 as the sampling operator, the sampling algorithm converges to $P(Q_i \mid \mathcal{E})$.*

Proof: It is sufficient to show that the expected value of $s_\mathcal{A}(q_i)$ is $P(\mathcal{E} \cup Q_i)$. The expected value is given by:

$$E[s(q_i)] = \sum_{\mathcal{A} \in \tilde{\mathcal{A}}} P_S(\mathcal{A}) s_\mathcal{A}(q_i) =$$

$$= \sum_{\mathcal{A} \in \tilde{\mathcal{A}}^{q_i}} P_S(\mathcal{A}) \frac{P(\mathcal{A})}{P_S(\mathcal{A})} + \sum_{\mathcal{A} \in \tilde{\mathcal{A}}^{q'}} P_S(\mathcal{A}) \frac{P(\mathcal{A})P(Q_i)}{P_S(\mathcal{A})}$$

and, by eliminating redundant factors, and using theorem 2, we have:

$$E[s(q_i)] = \sum_{\mathcal{A} \in \tilde{\mathcal{A}}^{q_i}} P(\mathcal{A}) + \sum_{\mathcal{A} \in \tilde{\mathcal{A}}^{q'}} P(\mathcal{A})P(Q_i) = P(\mathcal{E} \cup Q_i)$$

The theorem follows immediately. □

As for accumulation, the conditions of the theorem always hold for root nodes, and thus we can drop OPTION 1 from the sampling algorithm if we only need probability estimates for root nodes.

## 4 USING GENETIC ALGORITHMS

The fact that genetic algorithms (GA) [Goldberg, 1989] visit several points in the search space concurrently, appears useful for our purposes. At this point, we do not know how to define a sampling probability for the assignments generated by genetic algorithms. Thus, we use them, in this variant of the algorithm, only for computing bounds (accumulating), and not in the sampling probability estimator. Since we are not using samples, we can use either disjoint sets of hypercubes, or the sets of maximal hypercubes.

The difficulty with having a population consisting only of IB assignments is that such assignments may be incomplete. This implies that the number of individual nodes which have an assignment varies from one IB assignment to another, as well as which nodes are assigned. Traditional GA approaches assume population elements to be of a single fixed length.

Even if we permit incomplete assignments as population elements, we are not guaranteed that such assignments are IB. In fact, based on the IB condition from Definition 1, the strong constraints between individual node assignments renders nearly all incomplete assignments to be non-IB.

With the two problems of variable length elements and strong structural constraints, we consider a variant of GAs called messy GAs [Merkle and Lamont, 1993; Merkle and Lamont, 1994]. Intuitively, messy GAs employ a building-blocks approach for evolutionary programming. From a pool of genetic material consisting of both complete and incomplete genes, new genes are formed by cutting or splicing together existing genes. The better the genetic morsel, the more likely it will survive and help form new genes in each generation. The cutting and splicing operations effectively replace the crossover operation for GAs, but the mutation operation is still maintained.

For our problem, we choose hypercubes as our smallest genetic item. Our goal is to string these hypercubes together to form an IB assignment. Two problems must be accounted for. First, it is clear that certain hypercubes will be incompatible with others. Two such incompatible hypercubes in a gene should render it totally unfit. Second, we must somehow decide the fitness of these incomplete assignments. Obviously, these are the specific problems for this approach to working with IB assignments which we alluded to earlier.

The philosophy of messy GAs is to preserve/build "chunks" of genetic material, in this case, hypercube strings, which are very promising. Hence, hypercube strings which contain incompatibilities may be maintained until the desired "chunk" has been extracted or the offending substring is replaced. Thus, our fitness will be a function of the probabilities of the hypercubes involved merged with other factors such as compatibility and length.

## 5 EXPERIMENTAL RESULTS

Due to the hybrid nature of the algorithms, it is hard to get theoretical results on performance for any interesting class of problems, as was done in [Dagum and Chavez, 1993]. This is especially true since the kinds of problem instances we are working with have a high dependency value, and according to [Dagum and Chavez, 1993] are "expected" to be hard. We thus experimented on 2 problems, estimating probabilities of root nodes that are ancestors of the evidence nodes, as follows.

### 5.1 NETWORK FOR SENSOR FUSION

We experimented on a network, generated dynamically for fusion of sonar data in the presence of spurious readings, discussed in [Shimony and Berler, 1996]. These networks are essentially 3-level networks, where all evidence nodes are sink nodes, and we wish to compute the posteriors for all root nodes. The intermediate level consists of OR nodes. In the interest of keeping distribution arrays small, as well as decreasing the number of hypercubes per node, OR nodes were limited to 2 parents, by adding intermediate OR nodes where necessary. Evidence nodes were linked by a chain of AND nodes, so as to get only one actual evidence node. All nodes are binary valued. The network in the experiment had 21 original evidence nodes, 105 root nodes, for a total of 356 nodes (all relevant, being ancestors of some evidence node by construction). The network was expected to be extremely hard for randomized algorithms, since it has many conditional distribution entries of 0 (and thus the worst possible dependency value). Nevertheless, by performing sampling and discarding samples of with event-probability 0, sampling should still converge to correct values.

Comparisons were run using all the above algorithms, as well as the junction-tree exact algorithm variant in IDEAL. The network was near the limit of practical exact evaluation, taking 4.5 hours on a Sparc ELC, and nearly exhausting the swap space of 80MBytes. For the sampling algorithms, we ran 3000 samples each. For forward logic sampling and backward sampling, the network was extremely hard. Both algorithms generated a total of zero (0) useful samples, and approximation was thus impossible. The network was also hard for IB sampling, but we *did* get roughly 1% useful samples, to a reasonably fast, useful approximation, with an average (over the root nodes) error in posterior probability of 0.245 after 1000 samples, and 0.15 after 3000 samples.[4] This improvement in sampling results is due to both the cost-sharing heuristic, and the fact that the partial IB assignments left more possibilities open late in the sample generation process, and thus were less likely to run into being forced to select 0-probability terms than the other algorithms, which use only complete assignments. Trying to accumulate IB assignments proved useless. The very low evidence probability (about $10^{-14}$) made it impossible to collect sufficient non-evidence mass to achieve useful bounds, with the randomized algorithms (including GA), while the deterministic best-first search algorithm crashed, exhausting swap space, after several hours, before finding even the single most probable IB assignment consistent with the evidence.

---
[4] In the badly unoptimized current implementation, runtime was about 45 minutes (1 second per sample), but that can easily be improved by several orders of magnitude with simple programming techniques.

### 5.2 COMPARISON TO RELATED WORK

The second set consists of a 5-node network, with two sets of distributions. We show results for the first case in [Fung and Favero, 1994], for comparative purposes. The schemes tried were forward logic sampling and backward sampling (confirming experimental results by [Fung and Favero, 1994]), IB sampling (with $P_S$ based on cost-sharing), "optimal" sampling (that is, forward sampling according to the actual exact probability given the evidence), and IB sampling with accumulation of IB samples. Results of the first 4 algorithms are shown in figure 1. Errors are total errors for all variable states, averaged over 100-250 runs for the smaller sample counts, and 10 runs for the larger sample counts. The fact that IB sampling performed commensurately with "optimal" sampling is due to the fact that in this small network the cost-sharing heuristic approximates nearly exactly the prior distributions of all the nodes, which, admittedly, is a quirk of this network. [5] Similar results (not shown) occured for the modified network probabilities shown in [Fung and Favero, 1994]).

## 6  DISCUSSION

In the interest of finding the sampling probability $P_S$ for the assignments, we used sets of disjoint hypercubes at each node, rather than (possibly overlapping) maximal IB hypercubes. Using partial assignments to compute marginal probabilities has also been used by Poole [Poole, 1993a], where the partial assignments are disjoint *explanations* for the evidence, akin to our IB assignments covering the evidence. In fact, Poole's explanations *are* IB assignments, the only difference being that in our scheme there would be a somewhat smaller number of disjoint IB assignments, in many cases. Our results can thus be directly used as a sampling scheme for Poole's explanations.

One may ask how our experiments represent real (application) problems. A variant of this question is: How large a number of hypercubes are possible per Bayes net node? Is it about the same as the number of conditioning cases (bad), or the number of predecessors (good)? This depends on the structure of the node distribution. For pure OR, AND nodes, etc. the answer is favorable (linear in number of predecessors). The number is also small for noisy OR nodes, assuming that they are represented in causal independence [Heckerman and Breese, 1994] format: i.e. a pure OR with lead-in noise nodes. The latter can be done by a precompilation phase. Since many application Bayes nets have nodes of this type, and skewed distributions, we believe that our approximation algorithms will do well in many problem instances from applications.

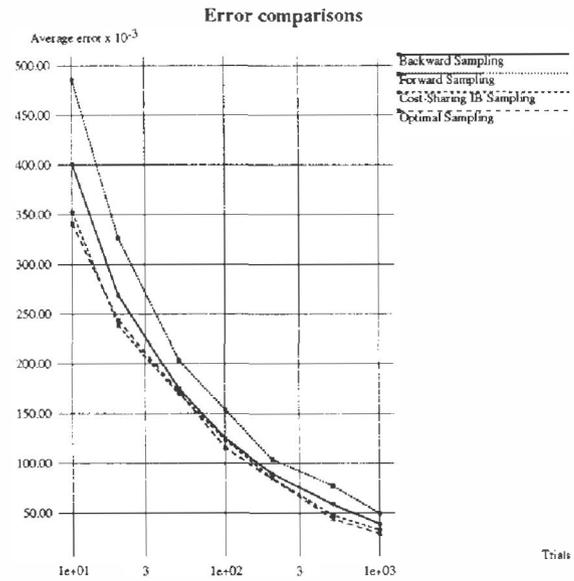

Figure 1: Comparison of Sampling Algorithms

## 7  CONCLUSION

Deterministic approximation algorithms for belief updating have the advantage of providing bounds on the probabilities, which are not available with sampling algorithms. Randomized algorithms have the advantage of providing approximations quickly, compared to the search performed in the deterministic algorithms, which may take exponential time, but provide no hard bounds.

This paper suggests a hybrid scheme: a randomized core that searches for good elements, and a deterministic accumulation of the probability mass in the elements, to get the hard bounds. In several (but certainly not all) cases, the hard bounds are of a magnitude similar to (or better) than the error estimates for sampling algorithms. Being more reliable, they provide a better probability estimate than sampling, in such cases. A novel variant of backward sampling, with sampled elements being partial IB assignments, rather than complete assignments, was also developed. Empirical evaluation of the algorithm showed its advantages over several existing sampling algorithms. Experimental results for IB sampling and accumulation clearly favor the sampling version of the algorithm over accumulation, for the sensor fusion network.

---

[5]The results for accumulated samples are an error margin $\Delta = 0$ for most runs. Presumably, this is due to the fact that the network is very small, and all possible states (cross product) were covered quickly. Needless to say, this is unlikely to occur in large networks.

Planned future work is to try to improve our GAs, for a better cover of the search space, and possibly to define a meaningful sample-probability for elements of a GA population. In the IB sampling algorithm, we intend to try to increase the fraction of useful IB samples, in networks for sensor fusion.

### Acknowledgements

This research was supported in part by AFOSR Project #940006, and by the Paul Ivanier center for robotics, Ben-Gurion University.

# References

[Bouckaert, 1994] Remco R. Bouckaert. A stratified simulation scheme for inference in Bayesian belief networks. In *Uncertainty in AI, Proceedings of the Tenth Conference*, pages 110–117, July 1994.

[Cooper, 1990] Gregory F. Cooper. The computational complexity of probabilistic inference using Bayesian belief networks. *Artificial Intelligence*, 42(2-3):393–405, 1990.

[Dagum and Chavez, 1993] P. Dagum and R. M. Chavez. Approximating probabilistic inference in bayesian belief networks. *Pattern Analysis and Machine Intelligence*, 15(3):244–255, March 1993.

[Dagum and Luby, 1993] Paul Dagum and Michael Luby. Approximating probabilistic inference in Bayesian belief networks is NP-hard. *Artificial Intelligence*, 60(1):141–153, 1993.

[D'Ambrosio, 1993] Bruce D'Ambrosio. Incremental probabilistic inference. In *Uncertainty in AI, Proceedings of the 9th Conference*, July 1993.

[Druzdzel, 1994] Marek Druzdzel. Some properties of joint probability distributions. In *Uncertainty in AI, Proceedings of the Tenth Conference*, pages 187–194, July 1994.

[Fung and Favero, 1994] Robert Fung and Brendan Del Favero. Backward simulation in Bayesian networks. In *Uncertainty in AI, Proceedings of the Tenth Conference*, pages 227–234, July 1994.

[Goldberg, 1989] David E. Goldberg. *Genetic Algorithms in Search, Optimization & Machine Learning*. Addison-Wesley Publishing Company, Inc., Reading, MA, 1989. Reprinted with corrections.

[Heckerman and Breese, 1994] David Heckerman and John S. Breese. A new look at causal independence. In *Uncertainty in AI, Proceedings of the Tenth Conference*, pages 286–292, July 1994.

[Henrion, 1988] Max Henrion. Propagating uncertainty in Bayesian networks by probabilistic logic sampling. In *Proceedings Uncertainty in Artificial Intelligence 4*, pages 149–163, 1988.

[Horvitz et al., 1989] Eric J. Horvitz, H. Jacques Suermondt, and Gregory F. Cooper. Bounded conditioning: Flexible inference for decisions under scarce resources. In *5th Workshop on Uncertainty in AI*, August 1989.

[Kjaerulff, 1994] Uffe Kjaerulff. Reduction of computational complexity in Bayesian networks through removal of weak dependencies. In *Uncertainty in AI, Proceedings of the Tenth Conference*, pages 374–382, July 1994.

[Li and D'Ambrosio, 1992] Z. Li and Bruce D'Ambrosio. An efficient approach to probabilistic inference in belief nets. In *Proceedings of the Annual Canadian AI Conference*, May 1992.

[Merkle and Lamont, 1993] Laurence D. Merkle and Gary B. Lamont. Comparison of parallel messy genetic algorithm data distribution strategies. In *Fifth International conference on Genetic Algorithms*, July 1993.

[Merkle and Lamont, 1994] Laurence D. Merkle and Gary B. Lamont. Scalability of parallel messy genetic algorithm data distribution strategies. In *Proceedings of the Symposium on Applied Computing Conference*, March 1994.

[Pearl, 1988] Judea Pearl. *Probabilistic Reasoning in Intelligent Systems: Networks of Plausible Inference*. Morgan Kaufmann, San Mateo, CA, 1988.

[Poole, 1993a] D. Poole. Probabilistic Horn abduction and Bayesian networks. *Artificial Intelligence*, 64(1):81–129, 1993.

[Poole, 1993b] David Poole. The use of conflicts in searching Bayesian networks. In *Uncertainty in AI, Proceedings of the 9th Conference*, July 1993.

[Santos and Shimony, 1994] Eugene Santos, Jr. and Solomon Eyal Shimony. Belief updating by enumerating high-probability independence-based assignments. In *Uncertainty in AI, Proceedings of the Tenth Conference*, pages 506–513, July 1994.

[Shimony and Berler, 1996] Solomon E. Shimony and Ami Berler. Bayes networks for sensor fusion in occupancy grids. Technical Report FC-96-01, Department of Math. and Computer Science, Ben-Gurion University, 1996.

[Shimony, 1994] Solomon E. Shimony. Finding MAPs for belief networks is NP-hard. *Artificial Intelligence Journal*, 68(2):399–410, August 1994.

[Wellman and Liu, 1994] Michael P. Wellman and Chao-Lin Liu. State-space abstraction for anytime evaluation of probabilistic networks. In *Uncertainty in AI, Proceedings of the Tenth Conference*, pages 567–574, July 1994.